\documentclass{article}
\usepackage{spconf,amsmath,graphicx,hyperref,booktabs,amssymb}
\usepackage{subcaption}
\usepackage{acro}
\usepackage[table]{xcolor} 
\usepackage[absolute,overlay]{textpos}
\DeclareAcronym{pttm}{short=PTTM, long={predictive turn-taking model}}
\DeclareAcronym{hri}{short=HRI, long={human-robot interaction}}
\DeclareAcronym{snr}{short=SNR, long={signal-to-noise ratio}}
\DeclareAcronym{asr}{short=ASR, long={automatic speech recognition}}
\DeclareAcronym{avsr}{short=AVSR, long={audio-visual speech recognition}}
\DeclareAcronym{vap}{short=VAP, long={voice-activity projection}}
\DeclareAcronym{mmvap}{short=MM-VAP, long={multimodal VAP}}
\DeclareAcronym{h/s}{short=H/S, long={hold/shift}}
\DeclareAcronym{ssl}{short={S3R}, long={self-supervised speech representation}}
\DeclareAcronym{wer}{short={WER}, long={word-error rate}}

\title{The Role of Prosodic and Lexical Cues in Turn-Taking with Self-Supervised Speech Representations}
\name{Sam O'Connor Russell\thanks{This work is financially supported by Taighde Éireann – Research Ireland under Grant No. 13/RC/2106 P2 and 22/FFP-A/11059}, Delphine Charuau and Naomi Harte}
\address{School of Engineering, Trinity College Dublin, Ireland}
\begin{document}

\maketitle

\begin{abstract}

Fluid turn-taking remains a key challenge in human-robot interaction. \Acp{ssl} have driven many advances, but it remains unclear whether \ac{ssl}-based turn-taking models rely on prosodic cues, lexical cues or both. We introduce a vocoder-based approach to control prosody and lexical cues in speech more cleanly than prior work. This allows us to probe the voice-activity projection model, an \ac{ssl}-based turn-taking model. We find that prediction on prosody-matched, unintelligible noise is similar to accuracy on clean speech. This reveals both prosodic and lexical cues support turn-taking, but either can be used in isolation. Hence, future models may only require prosody, providing privacy and potential performance benefits. When either prosodic or lexical information is disrupted, the model exploits the other without further training, indicating they are encoded in \acp{ssl} with limited interdependence. Results are consistent in CPC-based and wav2vec2.0 \acp{ssl}. We discuss our findings and highlight a number of directions for future work. All code is available to support future research. 

\end{abstract}


\section{Introduction}
\label{sec:intro}

\acresetall

Turn-taking is fundamental to human interaction, allowing conversations to flow smoothly \cite{levinson2015timing}. As speech planning takes longer than transitions between speakers, listeners predict when to speak ahead of time, supported by lexical and prosodic cues \cite{levinson2015timing}. Achieving fluid turn-taking is critical for human-robot interaction \cite{skantze2021turn}. However, most systems detect end-of-turn silence, leading to unnatural delays \cite{skantze2021turn}. 

Predictive turn-taking models address this by predicting speaker changes ahead of time \cite{ekstedt22_interspeech,roddy2018investigating,li2022can}. Feature engineering in earlier models, e.g. GeMAPS \cite{roddy2018investigating}, has since been replaced by \acp{ssl} \cite{ekstedt22_interspeech,li2022can}. \acp{ssl} encode rich phonetic, lexical and prosodic cues \cite{pasad2021layer,gimeno2025unveiling,vlasenko25_interspeech}, highly relevant to turn-taking \cite{levinson2015timing,cutler2018analysis}. Despite their efficacy, the richness of speech is collapsed into a single representation, making it challenging to establish what drives prediction. Prior work has shown \ac{ssl}-based turn-taking is quite robust to the flattening of pitch or intensity \cite{ekstedt22_interspeech}. Lexical information may therefore support prediction. Alternatively, as only pitch or intensity was flattened, the model may rely on the unmodified cue. Furthermore, we recently found background noise severely impacts turn-taking prediction \cite{oconnorrussell25_interspeech}. As noise did not fully degrade lexical content, prosodic and lexical cues in \ac{ssl} may be partially interdependent.  

These challenges reflect the wider issues around \ac{ssl} use in speech processing \cite{gimeno2025unveiling}. Their black-box nature hinders interpretability, limiting their ability to comply with ethics standards when decisions need to be explainable \cite{gimeno2025unveiling}. This has motivated a number of methods for studying \acp{ssl}. Probing provides a layer-wise analysis, but labels are required as small classifiers are trained with supervised learning \cite{pasad2021layer}. Input corruption can test sensitivity to a specific cue by removing it, without the need for labels \cite{ekstedt2022much}. A downside is common manipulations like low-pass filtering disrupt both intelligibility and prosody \cite{parsons25_interspeech}. Neuroscience routinely uses vocoders to cleanly isolate prosody while removing intelligibility, or vice versa \cite{brown2014evaluating,mushtaq2019evaluating}. To our knowledge, these techniques have not been applied to turn-taking or \ac{ssl} interpretability.

Motivated by the need for greater \ac{ssl} interpretability and recent work in turn-taking, we ask: (i) do \acp{ssl} support turn-taking primarily via prosodic cues, lexical cues, or both?; and (ii) are prosodic and lexical cues in \acp{ssl} interdependent? i.e., can one be disrupted whilst the other remains useful? 

Our contributions are threefold. Firstly, we introduce a vocoder method to cleanly control the prosodic and lexical information in speech. Secondly, we apply this method to turn-taking, showing that \ac{ssl} encodes both cues independently: when one is removed, the other is exploited without further training. Finally, we show that unintelligible noise matched to speech prosody achieves similar performance to clean speech. This motivates future prosody-only models, which have performance and privacy benefits. We conclude with a discussion of our findings for both \ac{ssl} and turn-taking. 

\section{Self-supervised speech representations and turn-taking}
\label{sec:background}

\acreset{ssl}

\Acp{ssl} have driven advances in speech processing as they capture rich representations of phonetics, prosody and semantics \cite{pasad2021layer,gimeno2025unveiling}; all of which are relevant to turn-taking \cite{levinson2015timing}. The \ac{vap} \cite{ekstedt22_interspeech} model is a state-of-the-art transformer-based turn-taking model leveraging \ac{ssl}. 
The \ac{vap} model is widely studied \cite{oconnorrussell25_interspeech,ekstedt2022much,inoue-etal-2024-multilingual}, though questions concerning its operation remain. Firstly, which cues drive prediction? An earlier study showed that when either pitch or intensity was flattened, performance fell but remained above baseline \cite{ekstedt2022much}. Thus, lexical cues might dominate prediction, but as only pitch or intensity was flattened, further work is needed. Secondly, it is unclear whether cues from speech captured by \acp{ssl} are interdependent. Our recent study suggests interdependence: light noise, which disrupts prosody but leaves speech mostly intelligible, leads to poor prediction \cite{oconnorrussell25_interspeech}. 

Several methods to better understand \ac{ssl} models exist. Probing provides a layer-wise analysis \cite{pasad2021layer}. Input corruption directly manipulates speech to understand how a cue is captured by \acp{ssl}. Common methods such as low-pass filtering impact both intelligibility and prosody, limiting insights \cite{ekstedt22_interspeech,parsons25_interspeech}. Methods which more cleanly isolate cues are used in neuroscience to study language processing \cite{brown2014evaluating,mushtaq2019evaluating} akin to the highly entangled nature of \acp{ssl}. Vocoders decompose speech into spectral and amplitude components, enabling isolation of prosodic and lexical cues \cite{brown2014evaluating}. Yet to our knowledge, they have not been applied to turn-taking with \ac{ssl}. 

\section{Isolating Prosodic and Lexical Cues} 
\label{sec:method}

We explore \ac{ssl}-based turn-taking with the CANDOR corpus \cite{reece2023candor} used in previous work \cite{oconnorrussell25_interspeech}. It consists of 1,657 US English dyads (850 hrs). Audio is 16kHz stereo (1 per speaker) and we used Amazon Transcribe for timing information.

\begin{figure}
    \centering
    \includegraphics[width=\columnwidth]{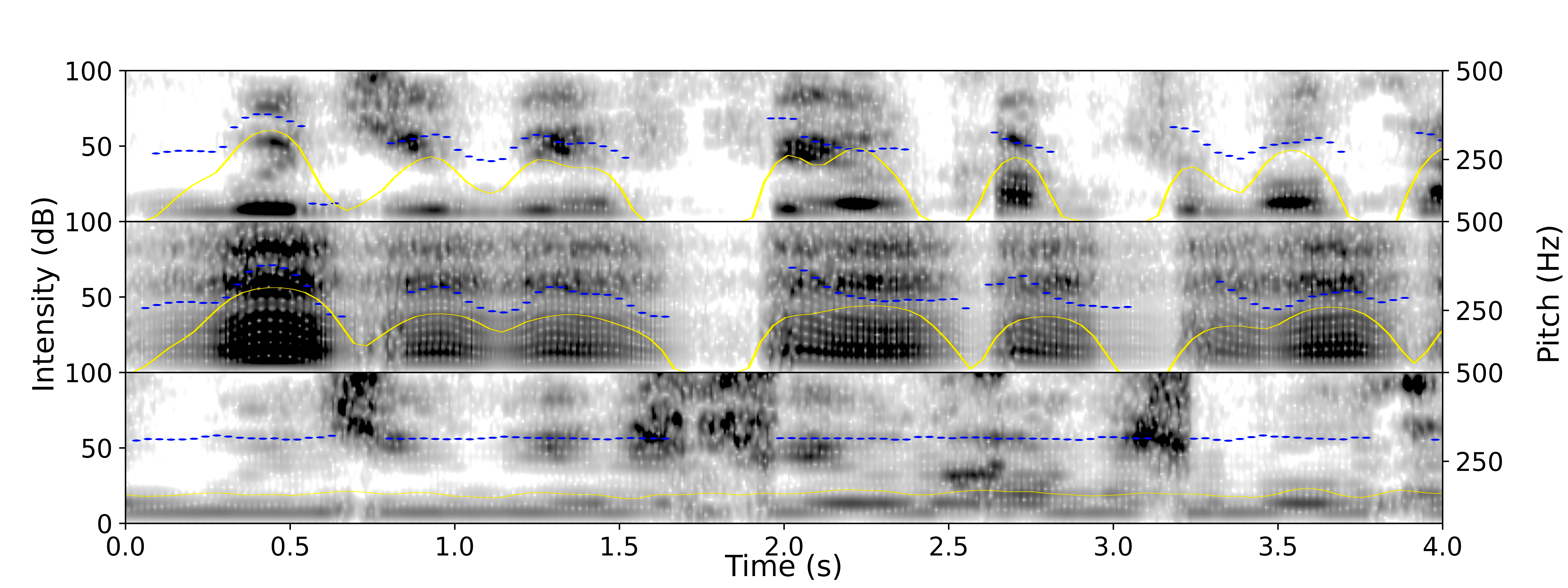}
    \caption{Spectrograms utterance ``would you feel comfortable". Top=original, middle=noise+prosody preserved, bottom=intelligible+prosody flat. Pitch=blue, intensity=yellow.}
    \label{fig:spectrogram}
\end{figure}

To control prosodic and lexical cues, we use the \textbf{WORLD vocoder} \cite{morise2016world}\footnote{30 s windows, 512 ffts, frame period 10 ms}. We generate \textbf{prosody-matched noise} by replacing the spectral envelope with pink noise (removing intelligibility), preserving pitch and amplitude. This isolates prosody by removing lexical information. We further isolate pitch and intensity in two additional noises, which flatten either the pitch or intensity to the utterance mean. We conduct \textbf{prosodic manipulation} preserving the spectral envelope (intelligibility) and flattening pitch, intensity, or both. A spectrogram of a sample utterance  is in Figure \ref{fig:spectrogram}, showing the absence (middle) and preservation (bottom) of spectral information. We also generate \textbf{background noise} (babble, music and speech=random utterances) by following our prior work \cite{oconnorrussell25_interspeech}. 

We add noise to speech at controlled \acp{snr} (-10:10 dB, step 2.5) \cite{oconnorrussell25_interspeech}. We compute the \ac{wer} with Whisper transcription \cite{radford2023robust} to quantify the removal of lexical information, a common proxy for intelligibility e.g. \cite{choi23_interspeech}. Samples and code are available\footnote{\url{github.com/russelsa/noise_generation_ICASSP-}}. 

We use the \textbf{VAP turn-taking model} which predicts the next 2 seconds of speaking activity (20Hz) \cite{ekstedt22_interspeech,inoue-etal-2024-multilingual}. The front-end is a pre-trained \ac{ssl}  \cite{riviere2020unsupervised} (causal, frozen). The model has 3 self- (1 per channel, tied weights) and 1 cross-attention transformer layers (256d) \cite{riviere2020unsupervised}. We withhold 70 test sessions and cross validate on remaining sessions (5 folds). We train 10 VAP models (Table \ref{tab:models}) where 'mixed' demotes 25\% of training sessions with 0 dB of noise / prosodic manipulation \cite{oconnorrussell25_interspeech}. We train each for 10 epochs (lr=1e-4, batch=32) on an RTX 6000 GPU. We train a \ac{vap} variant with a wav2vec2.0 \cite{baevski2020wav2vec} front-end (projected to 256d). 

\begin{table}[h]
\centering
\caption{Overview of the 10 VAP model training conditions.}
\resizebox{\linewidth}{!}{
\begin{tabular}{lll}
\hline
\textbf{Model description} & \textbf{Model training data} & \textbf{\# Models} \\
\hline
Clean speech & Clean speech & 1 \\
Prosody-matched noise (no lexical cues) & Noise matching pitch, intensity, and pitch+intensity  & 3 \\
Prosodic manipulation & Speech with flattened pitch, intensity and pitch+intensity & 3 \\
Mixed & 75\% clean + 25\% noise matched to prosody & 1 \\
Mixed & 75\% clean + 25\% prosody flattened speech & 1 \\
Mixed & 75\% clean + 25\% background noise & 1 \\
\hline
\end{tabular}}
\label{tab:models}
\end{table} 


We evaluate models by extracting shifts (speaker change after $>$200\,ms silence; 83,158) and holds (no speaker change; 206,830) following \cite{ekstedt22_interspeech}. \textbf{S/H-Pred} is a binary comparison of speech before shifts with speech before holds, whereas \textbf{S-Pred} compares pre-shift speech with mid-turn speech (83,158). We sum model probabilities over 200 ms, applying a binary threshold (tuned on the validation set). We report test set F1 and balanced accuracy with t-tests for significance.


\begin{figure*}[!htbp]
    \centering
    \begin{subfigure}[b]{0.35\textwidth}
        \centering
        \includegraphics[width=\textwidth]{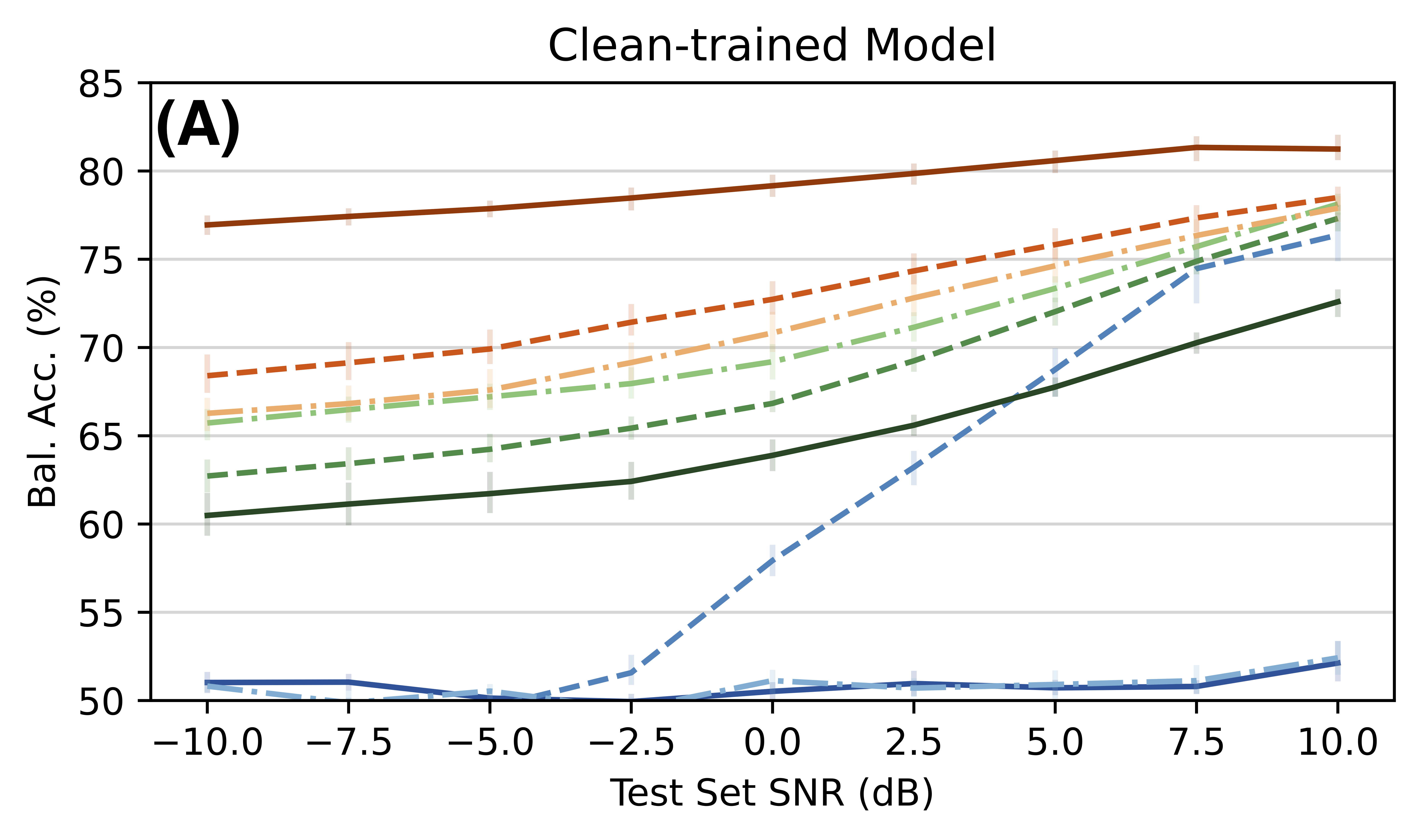}
        \label{fig:noaug-snr-sweep}
    \end{subfigure}
    \hspace{2mm}
    \begin{subfigure}[b]{0.35\textwidth}
        \centering
        \includegraphics[width=\textwidth]{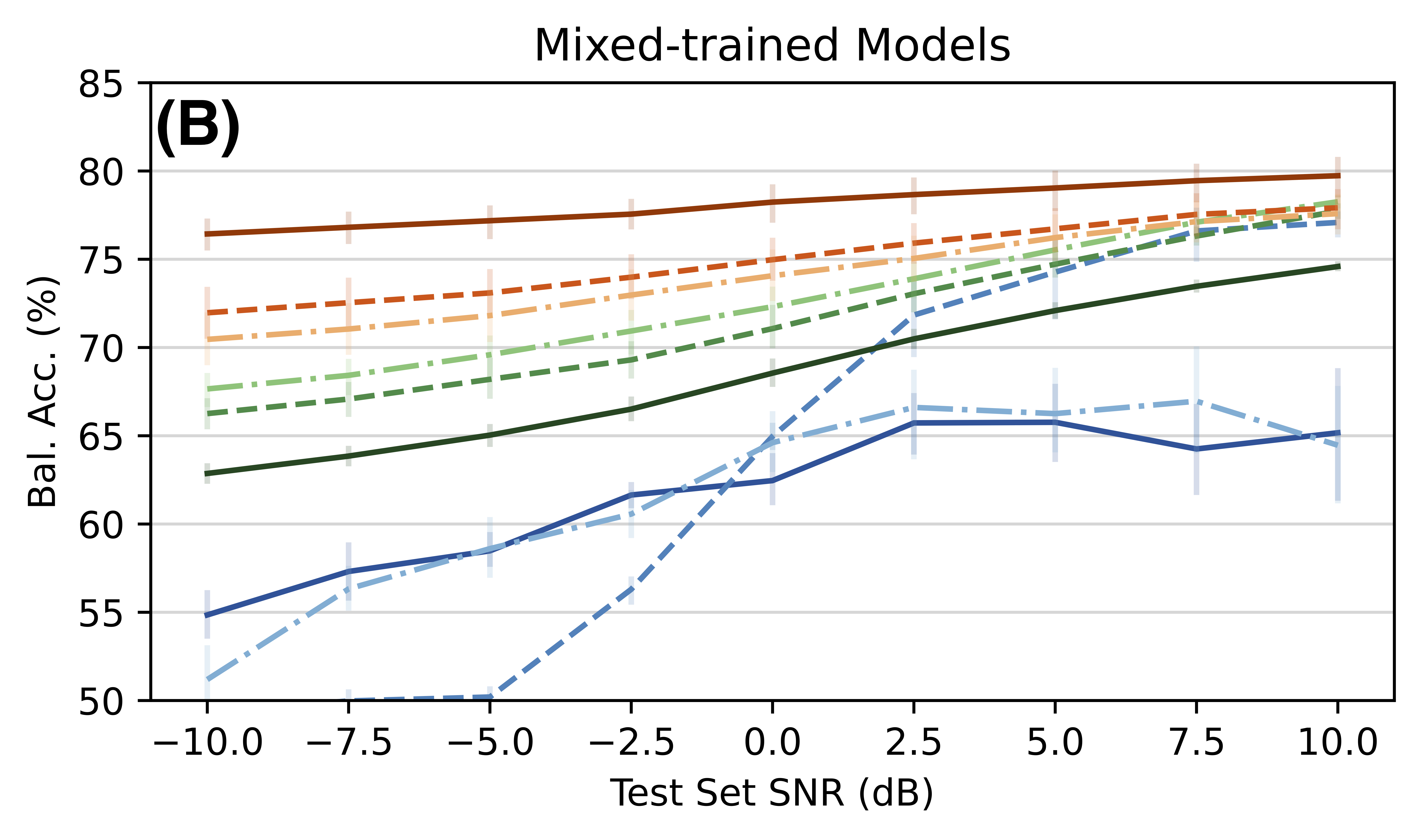}
        \label{fig:aug-snr-sweep}
    \end{subfigure}
    \hspace{2mm}
    \begin{subfigure}[b]{0.20\textwidth}
        \centering
        \raisebox{1.0cm}{%
        \includegraphics[width=\textwidth]{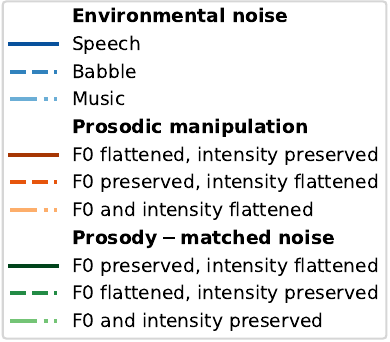}
        }
    \end{subfigure}

    \caption{VAP model performance (S/H-Pred metric, 5-fold average $\pm 95\%$ c.i.) trained on clean speech (A) and a mixture of clean and manipulated speech (B). Balanced accuracy for each speech manipulation by test set signal-to-noise ratio (SNR).}
    \label{fig:snr-sweep-accuracy}
\end{figure*}

\section{Exploring self-supervised representations in turn-taking}
\label{sec:results}
\subsection{Turn-taking in clean and manipulated speech}

We first train and test the VAP model on clean speech (Table \ref{tab:clean-speech}). We note strong performance in line with prior work (80 and 85\% balanced accuracy for S/H-Pred and S-Pred) \cite{oconnorrussell25_interspeech}. When pitch and intensity are flattened, performance drops but remains above chance (e.g. $85\rightarrow68\%$ for S-Pred). The model is more sensitive to flattening of intensity than pitch. Prosodic noise removes lexical information whilst preserving prosody. Testing on prosodic noise yields a larger drop ($85\rightarrow60\%$), though performance remains above chance.

Overall, the model is more robust to the removal of prosodic information than lexical. Performance drops when either cue is removed but above-chance performance demonstrates the model can exploit whichever cue remains available. 

\begin{table}[!htbp]
\centering
\caption{VAP model performance on clean speech and on acoustic manipulations, grouped by their preservation of lexical content and prosody (5-fold average, $\pm$ std \checkmark=preserved, x=flattened. \checkmark \checkmark \checkmark = original speech. $F_1$ (w)=weighted).}
\label{tab:clean-speech}
\resizebox{0.95\columnwidth}{!}{%
\begin{tabular}{lccccccc}
\toprule
\textbf{Metric} & \multicolumn{3}{c}{\textbf{Test set contains}} & $\mathbf{F_1}$ \textbf{(w)} & $\mathbf{F_1}$ \textbf{(Hold)} & $\mathbf{F_1}$ \textbf{(Shift)} & \textbf{Bal. Acc. (\%)} \\ 
 & \textbf{Lexical} & \textbf{Pitch} & \textbf{Intensity} &  & & &  \\ 
\midrule
\textit{S-Pred} & \cellcolor{green!15}\checkmark & \cellcolor{green!15}\checkmark & \cellcolor{green!15}\checkmark & 0.86 & 0.84 & 0.87 & 85 ± 0 \\ \cline{2-8}
 & \cellcolor{red!15}x & \cellcolor{red!15}x & \cellcolor{green!15}\checkmark & 0.58 & 0.54 & 0.62 & 58 ± 2 \\ 
 & \cellcolor{red!15}x & \cellcolor{green!15}\checkmark & \cellcolor{red!15}x & 0.58 & 0.52 & 0.64 & 58 ± 2 \\
 & \cellcolor{red!15}x & \cellcolor{green!15}\checkmark & \cellcolor{green!15}\checkmark & 0.60 & 0.56 & 0.64 & 60 ± 2 \\ \cline{2-8}
 & \cellcolor{green!15}\checkmark & \cellcolor{red!15}x & \cellcolor{green!15}\checkmark & 0.76 & 0.74 & 0.80 & 76 ± 1 \\
 & \cellcolor{green!15}\checkmark & \cellcolor{green!15}\checkmark & \cellcolor{red!15}x & 0.71 & 0.67 & 0.74 & 71 ± 1 \\
 & \cellcolor{green!15}\checkmark & \cellcolor{red!15}x & \cellcolor{red!15}x & 0.68 & 0.63 & 0.72 & 68 ± 1 \\ \midrule
\textit{S/H-Pred} & \cellcolor{green!15}\checkmark & \cellcolor{green!15}\checkmark & \cellcolor{green!15}\checkmark & 0.83 & 0.88 & 0.70 & 80 ± 1 \\ \cline{2-8}
 & \cellcolor{red!15}x & \cellcolor{red!15}x & \cellcolor{green!15}\checkmark & 0.67 & 0.81 & 0.33 & 59 ± 1 \\
 & \cellcolor{red!15}x & \cellcolor{green!15}\checkmark & \cellcolor{red!15}x & 0.66 & 0.80 & 0.31 & 57 ± 2 \\
 & \cellcolor{red!15}x & \cellcolor{green!15}\checkmark & \cellcolor{green!15}\checkmark & 0.69 & 0.81 & 0.37 & 61 ± 1 \\ \cline{2-8}
 & \cellcolor{green!15}\checkmark & \cellcolor{red!15}x & \cellcolor{green!15}\checkmark & 0.76 & 0.84 & 0.56 & 72 ± 1 \\
 & \cellcolor{green!15}\checkmark & \cellcolor{green!15}\checkmark & \cellcolor{red!15}x & 0.72 & 0.81 & 0.46 & 66 ± 1 \\
 & \cellcolor{green!15}\checkmark & \cellcolor{red!15}x & \cellcolor{red!15}x & 0.70 & 0.80 & 0.43 & 63 ± 1 \\ \bottomrule
\end{tabular}%
}
\end{table}

\subsection{Effect of prosodic manipulation and noise}

\begin{figure}[!htbp]
    \centering
    \includegraphics[width=0.95\columnwidth]{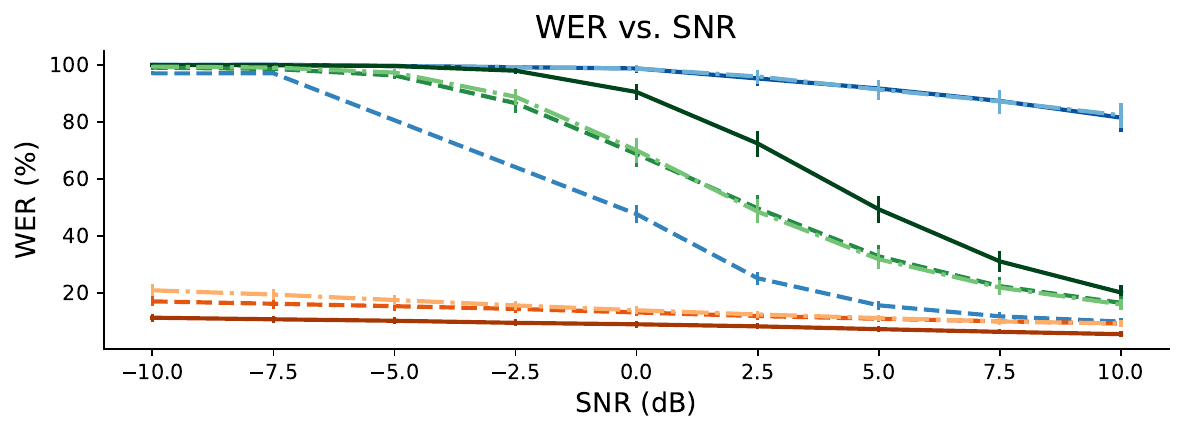}
    \caption{Word error rate for each manipulation (WER, clamped to 100\%) at each SNR in decibels (dB). P=pitch, I=intensity. Mean per speaker $\pm 95\%$ c.i. Legend as above.}
    \label{fig:wer-snr}
\end{figure}

Next, we explore how removing prosodic/lexical manipulations impacts intelligibility. Figure \ref{fig:wer-snr} shows the WER vs SNR. Additive prosody-matched noise removes lexical content (in green, WER $\rightarrow100\%$), whereas prosodic manipulation preserves lexical cues (in orange, WER $<20\%$). Background noise degrades lexical content as the SNR decreases (blue). 

We then test the clean speech-trained VAP model on speech with prosodic/lexical manipulation. Figure \ref{fig:snr-sweep-accuracy} A shows S/H-pred accuracy vs test set SNR\footnote{S-pred omitted for space, clean speech omitted for clarity (Table \ref{tab:clean-speech})}. At high SNR, flattening of prosody has no significant impact on accuracy ($80\%$ vs $80\%$ $ p>0.05$) except when the intensity contour is flattened ($80\%\rightarrow73\%$). Although no lexical information is preserved, at -10dB SNR accuracy is 66\% in prosody-matched noise (WER $>$ 100\%). Background noise (blue) has similarly high WER yet accuracy is 52\%. Thus, these results support prosody as an independent turn-taking cue. 

We then train models on a mix of clean and distorted speech with results in Figure \ref{fig:snr-sweep-accuracy} B. The increased performance shows the model exploits prosodic and lexical cues more effectively (orange and green). Background noise performance did not improve in low SNR (blue), where lexical information is removed. At high SNR there is some increased performance, yet it is lower than the performance of other models. Thus either prosodic or lexical information must be preserved for robust performance. When one is removed, the other can be flexibly used by the model. This shows that prosodic and lexical cues are not interdependent in \acp{ssl}. 

\subsection{Prosody and lexical cues in turn-taking prediction}

\begin{figure}[!h]
    \centering
    \includegraphics[width=\columnwidth]{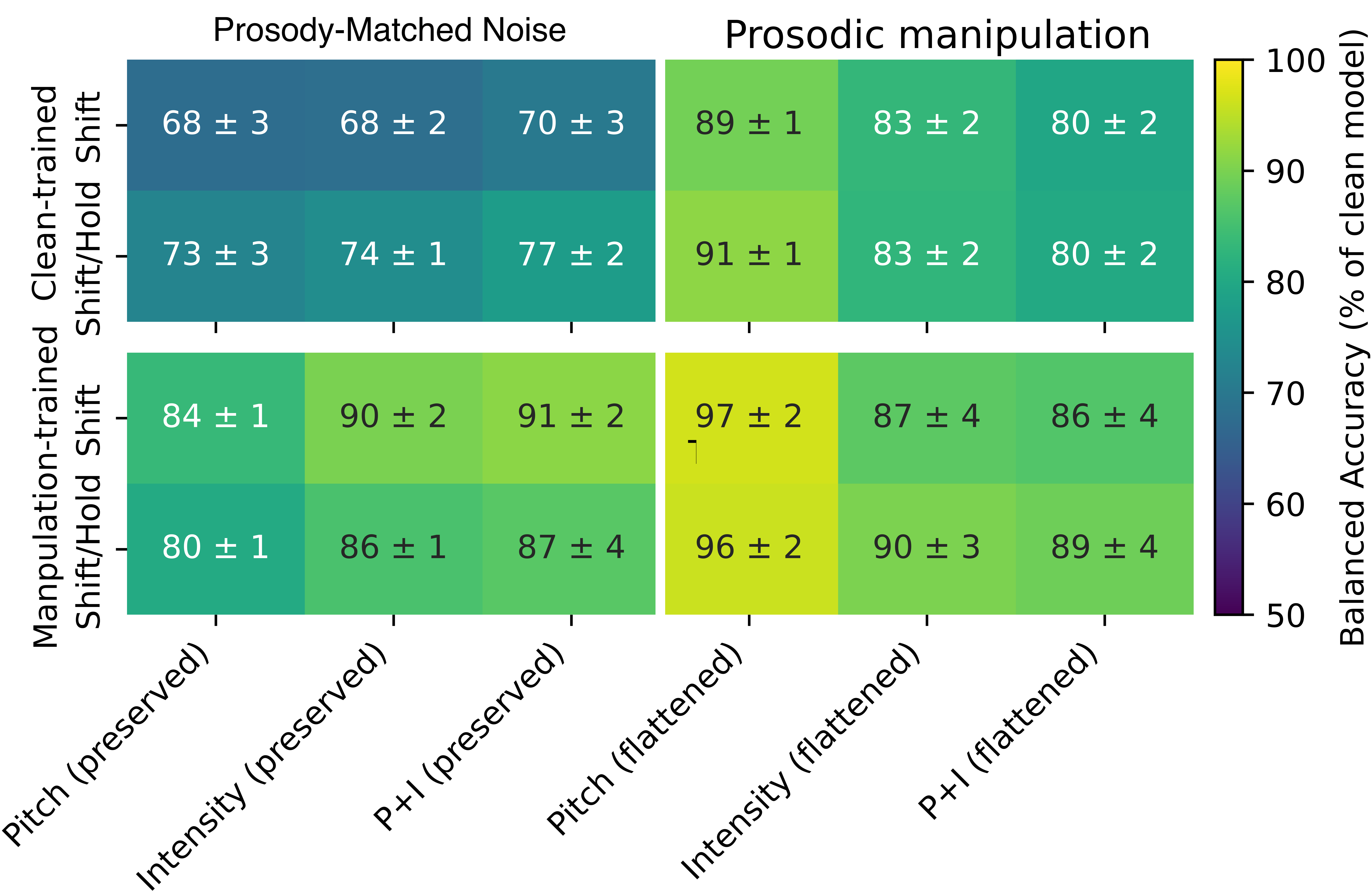}
    \caption{Percentage of clean speech model accuracy achieved by when tested (top) and trained (bottom) on prosody-matched noise and speech with flat prosody. 
    }
    \label{fig:heatmap}
\end{figure}

We next train models exclusively on manipulated speech (Table \ref{tab:noise-trained}). We express results as a percentage of the accuracy of the model trained and tested on clean speech in Figure \ref{fig:heatmap}, establishing to what extent prosody and lexical cues are independent predictors. Noise matching the original prosody achieves 87-91\% of clean accuracy, comparable to models trained on prosody flattened speech (89-96\%). Hence isolating prosodic and lexical cues yields accuracy close to clean speech, supporting these cues as independent predictors. 

\begin{table}[!h]
\centering
\caption{VAP model performance, trained and tested on either clean speech or speech manipulated speech (5-fold average $\pm$ std. \checkmark=preserved, x=flattened. $F_1$ (w)=weighted).}
\label{tab:noise-trained}
\resizebox{0.95\columnwidth}{!}{%
\begin{tabular}{lccccccc}
\toprule
\textbf{Metric} & \multicolumn{3}{c}{\textbf{Test + train sets contain}} & $\mathbf{F_1}$ \textbf{(w)} & $\mathbf{F_1}$ \textbf{(Hold)} & $\mathbf{F_1}$ 
\textbf{(Shift)} & \textbf{Bal. Acc. (\%)} \\ 
 & \textbf{Lexical} & \textbf{Pitch} & \textbf{Intensity} &  & & &  \\ 
\midrule
\textit{S-Pred} & \cellcolor{green!15}\checkmark & \cellcolor{green!15}\checkmark & \cellcolor{green!15}\checkmark & 0.86 & 0.84 & 0.80 & 85 ± 0.4 \\ \cline{2-8}
 & \cellcolor{red!15}x & \cellcolor{red!15}x & \cellcolor{green!15}\checkmark & 0.77 & 0.74 & 0.75 & 77 ± 1.7 \\
 & \cellcolor{red!15}x & \cellcolor{green!15}\checkmark & \cellcolor{red!15}x & 0.72 & 0.68 & 0.81 & 71 ± 0.8 \\ 
 & \cellcolor{red!15}x & \cellcolor{green!15}\checkmark & \cellcolor{green!15}\checkmark & 0.78 & 0.75 & 0.84 & 78 ± 1.5 \\  \cline{2-8}
 & \cellcolor{green!15}\checkmark & \cellcolor{red!15}x & \cellcolor{green!15}\checkmark & 0.82 & 0.80 & 0.83 & 82 ± 1.8 \\
 & \cellcolor{green!15}\checkmark & \cellcolor{green!15}\checkmark & \cellcolor{red!15}x & 0.81 & 0.78 & 0.83 & 80 ± 1.5 \\
 & \cellcolor{green!15}\checkmark & \cellcolor{red!15}x & \cellcolor{red!15}x & 0.74 & 0.71 & 0.70 & 74 ± 3.4 \\
 \midrule
\textit{S/H-Pred} & \cellcolor{green!15}\checkmark & \cellcolor{green!15}\checkmark & \cellcolor{green!15}\checkmark & 0.83 & 0.88 & 0.70 & 80 ± 1 \\ \cline{2-8}
 & \cellcolor{red!15}x & \cellcolor{red!15}x & \cellcolor{green!15}\checkmark & 0.72 & 0.82 & 0.43 & 68 ± 1.2 \\ 
 & \cellcolor{red!15}x & \cellcolor{green!15}\checkmark & \cellcolor{red!15}x & 0.70 & 0.81 & 0.49 & 64 ± 1.0 \\
 & \cellcolor{red!15}x & \cellcolor{green!15}\checkmark & \cellcolor{green!15}\checkmark & 0.73 & 0.82 & 0.62 & 69 ± 3.2 \\ \cline{2-8}
 & \cellcolor{green!15}\checkmark & \cellcolor{red!15}x & \cellcolor{green!15}\checkmark & 0.82 & 0.85 & 0.55 & 76 ± 1.4 \\ 
 & \cellcolor{green!15}\checkmark & \cellcolor{green!15}\checkmark & \cellcolor{red!15}x &  0.75 & 0.83 & 0.53 & 71 ± 2.4 \\
 & \cellcolor{green!15}\checkmark & \cellcolor{red!15}x & \cellcolor{red!15}x & 0.75 & 0.83 & 0.52 & 70 ± 3.0 \\ 
 \bottomrule
\end{tabular}%
}
\end{table}
 
\subsection{Generalisation to wav2vec2.0}
We train a variant with wav2vec2.0 encoder \cite{baevski2020wav2vec} finding prosodic or lexical cues can be removed with minimal impact on performance ($93\%$ vs $92\%$). Unlike the earlier models and human turn-taking, wav2vec2.0 embeddings contain information from future frames, leading to unrealistic performance. Results affirm prior findings are not limited to a single \ac{ssl}. 

\begin{table}[!h]
\centering
\caption{Wav2vec2.0 trained on clean speech (5-fold average $\pm$ std. \checkmark=preserved, x=flattened. $F_1$ (w)=weighted).}
\label{tab:wav2vec}
\resizebox{0.95\columnwidth}{!}{%
\begin{tabular}{lccccccc}
\toprule
\textbf{Metric} & \multicolumn{3}{c}{\textbf{Test sets contain}} & $\mathbf{F_1}$ \textbf{(w)} & $\mathbf{F_1}$ \textbf{(Hold)} & $\mathbf{F_1}$ 
\textbf{(Shift)} & \textbf{Bal. Acc. (\%)} \\ 
 & \textbf{Lexical} & \textbf{Pitch} & \textbf{Intensity} &  & & &  \\ 
\midrule
\textit{Shift/hold} & \cellcolor{green!15}\checkmark & \cellcolor{green!15}\checkmark & \cellcolor{green!15}\checkmark & 0.94 & 0.95 & 0.89 & 93 ± 0  \\ 
 & \cellcolor{red!15}x & \cellcolor{green!15}\checkmark & \cellcolor{green!15}\checkmark & 0.92 & 0.94 & 0.85 & 92 ± 1 \\ 
 & \cellcolor{green!15}\checkmark & \cellcolor{red!15}x & \cellcolor{red!15}x & 0.92 & 0.94 & 0.85 & 92 ± 1 \\ 
 \bottomrule
\end{tabular}%
}
\end{table}

\section{Discussion of Key Findings}

\acreset{ssl}
Turn-taking remains a key unsolved problem in human-robot interaction. \Acp{ssl} have transformed turn-taking and speech processing, though interpretability remains challenging. In this paper, we take a new approach to analysis, isolating prosodic and lexical information in speech. 
We showed a VAP turn-taking model trained on noise matched to original prosody retains 91\% of the accuracy of a model trained and tested on clean speech. Thus prosody alone can function as a turn-taking cue. Similarly, speech with flat prosody and just lexical cues supports turn-taking with high accuracy. Performance is still better when both prosodic and lexical cues are present (clean speech). To answer of our first research question, both prosody and lexical cues can function as independent turn-taking predictors, and both support prediction. 

Our second question explored whether prosodic and lexical cues are interdependent in \acp{ssl}. Performance is remarkably similar in unintelligible speech matched to the original prosody ($\approx70\%$), even when the model was trained on clean speech ($\approx66\%$). Hence, an \ac{ssl}-based turn-taking model can rely on prosody when lexical information is degraded and vice-versa. These cues are therefore not interdependent in \ac{ssl}. We showed this was also true for a wav2vec2.0 model.

Our work has revealed much about \ac{ssl}-based turn-taking and \acp{ssl}, raising a number of directions for future work. Our method isolates these cues more cleanly than prior work \cite{parsons25_interspeech}. It should therefore be used to clarify the extent to which \acp{ssl} exploit lexical information in outperforming engineered prosodic features, not addressed in prior work \cite{kakouros25b_interspeech}. Secondly, based on our finding that prosody alone supports turn-taking, lightweight prosody-based models should be considered. Another advantage is that as we have shown, isolating prosody removes useful lexical information. This has significant privacy advantages e.g. sensitive topics removed. Furthermore, as both the spectral envelope and pitch contours display individual speaker differences \cite{akagi1997speaker}, using the prosodic contour reduces, but does not fully eliminate, identifiable information. Finally, future work should explore cross-linguistic generalisation of \acp{ssl} with our method, building on prior work \cite{inoue-etal-2024-multilingual}. 


A limitation is the potential for residual phonetic information to persist in prosody-matched noise, though this is unlikely, as all spectral information is removed. Furthermore, the WER value confirms no useful information persists. Finally, we only consider English in the CANDOR corpus and future work should consider other corpora. However, this corpus is large (850 hr) and representative, containing many speakers across $\approx$1600 dyads of casual talk.

\section{Conclusion}
\acreset{ssl}
We revealed that \ac{ssl} independently encode prosodic and lexical cues: when one is disrupted, \ac{ssl}-based turn-taking models reliably use the other without additional training. We further showed that the prosodic contour is sufficient for turn-taking, with performance on noise matched to prosody nearly matching clean speech performance. Our findings enhance \ac{ssl} interpretability, motivating future work in prosody-only turn-taking models, which have important advantages including privacy.  


\bibliographystyle{IEEEbib}
\bibliography{strings,refs}

\end{document}